\documentclass{article}

\PassOptionsToPackage{numbers}{natbib}
\usepackage[preprint]{neurips_2026}
\usepackage[utf8]{inputenc} 
\usepackage[T1]{fontenc}    
\usepackage{hyperref}       
\usepackage{url}            
\usepackage{booktabs}       
\usepackage{amsfonts}       
\usepackage{nicefrac}       
\usepackage{microtype}      
\usepackage{xcolor}         
\usepackage{graphicx}
\usepackage{multirow}
\usepackage{amsmath}
\usepackage{placeins}


\title{Source-Anchored Visual Autoregression for Extreme Image Compression}

\author{%
  Bingxin Wang\\
  Department of Mathematics, HKUST\\
  \texttt{bwangbo@connect.ust.hk}
  \And
  Yuan Lan\\
  Huawei Hong Kong Research Center\\
  \texttt{ylanaa@connect.ust.hk}
  \And
  Yang Xiang\thanks{Corresponding author.}\\
  Department of Mathematics, HKUST\\
  \texttt{maxiang@ust.hk}
  \AND
  Yiqing Huang\\
  Tsinghua University
}

\begin{document}

\maketitle

\begin{abstract}
In the extreme-low bitrate range below $0.01$ bpp, image compression becomes a constrained generation problem: the reconstruction depends on inferring plausible missing details while staying close to the input image. Autoregressive token models provide an efficient coarse-to-fine prior for this setting, but the transmitted early-scale tokens often leave object identity and texture underdetermined, which can lead to source-inconsistent reconstructions. We introduce \textbf{Source-Anchored Visual Autoregression} (SAVAR), a generative codec built on Visual Autoregressive (VAR) modeling. SAVAR transmits early-scale tokens and a compact caption. The tokens provide image-specific layout information, while the caption supplies global semantics. SAVAR reconstructs a partial VQ-VAE latent from the retained scales, maps it to the start-of-sequence initialization, and uses the same latent to condition the caption embedding before adaptive layer normalization, separating two roles that are typically driven by a shared condition in VAR-style generation. On the Kodak and CLIC2020 datasets in the extreme-low range, SAVAR reduces perceptual distortion relative to matched ARPC retained-scale operating points, lowering LPIPS by about $8$--$9\%$ and DISTS by up to $22\%$ at a lower bpp. It reconstructs a $512 \times 512$ image in about $218$ ms on one NVIDIA L20X GPU, retaining fast autoregressive decoding in the evaluated setting.
\end{abstract}

\section{Introduction}
\label{sec:intro}

Extreme image compression requires a decoder to recover a recognizable image from only a few thousand bits. Below $0.01$ bpp, the bitstream usually cannot specify a single high-resolution reconstruction: much of the texture, object detail, and sometimes even semantic identity has to be inferred. This makes the task different from conventional rate-distortion optimization. A codec therefore benefits from a generative prior for synthesizing missing content, while the reconstruction should remain close to the input image. GAN-based codecs \cite{mentzer2020high, Agustsson_2019_ICCV} and diffusion-based codecs \cite{yang2024lossy, careil2023towards, xue2025dlf} improve perceptual quality under this rate-distortion-perception tradeoff \cite{blau2019rethinking}, but diffusion models typically require iterative denoising and therefore introduce higher decoding latency.

Autoregressive (AR) models over discrete visual tokens \cite{van2017neural, esser2021taming, chang2022maskgit} can provide a faster generative route for extreme-low bitrate compression. VAR modeling \cite{xue2024visual} predicts visual tokens scale by scale, providing a natural coarse-to-fine generation process. ARPC \cite{zhang2026arpc} turns this property into a progressive image codec: it transmits the first few scale tokens and uses the autoregressive model to generate the missing higher-resolution scales. This design is attractive because rate control becomes scale selection, and decoding avoids a long diffusion chain.

However, progressive transmission alone does not remove all ambiguity. The early-scale tokens mainly carry coarse layout and low-frequency appearance. They may not determine object identity, fine structure, or texture, so the AR prior fills in the missing content. A caption can provide high-level semantics, but it often describes a set of plausible images rather than uniquely specifying the source image to be reconstructed. The result may look realistic while altering the source content. We refer to this source-inconsistent completion as \emph{semantic drift}. As shown by the path-level ablation in Table~\ref{tab:diagnosis}, replacing either the start-of-sequence ($\mathrm{SOS}$) initialization from transmitted tokens or the retained-scale-aware adaptive layer normalization (AdaLN) conditioning with text-only control degrades perceptual and structural fidelity under the same bitrate.

\begin{figure}[!t]
    \centering
    \includegraphics[page=1,width=\textwidth]{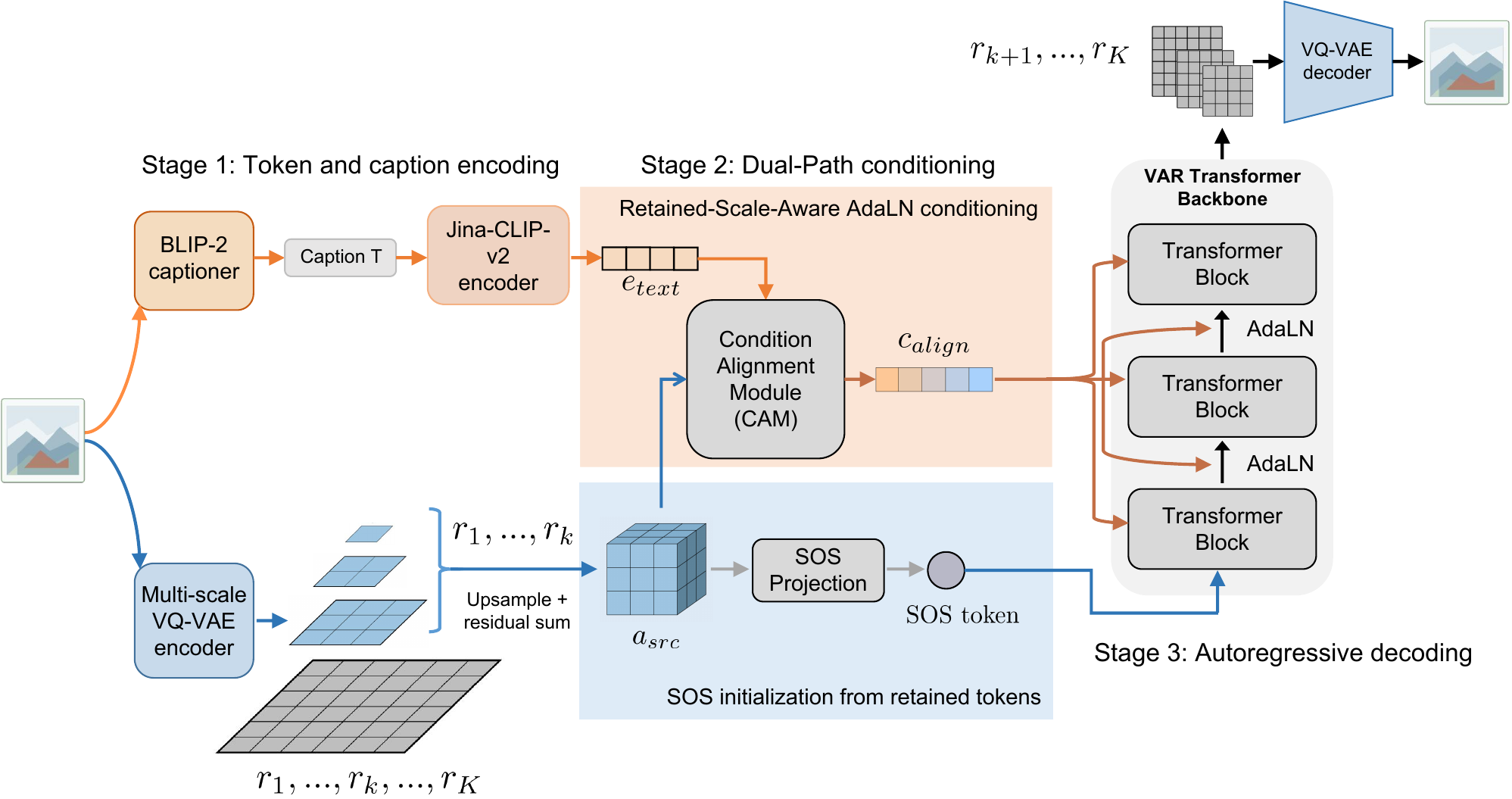}\par
    \vspace{-0.65em}
    {\footnotesize\sffamily\textbf{(a) SAVAR overview}\par}
    \vspace{-3.2em}
    \includegraphics[page=2,viewport=0 20 960 275,clip,width=\textwidth]{pic1.pdf}\par
    \vspace{0.8em}
    {\footnotesize\sffamily\textbf{(b) CAM for retained-scale-aware text conditioning}\par}
    \vspace{0.2em}
    \caption{Overview of SAVAR. (a) The retained scale tokens are mapped to a partial VQ-VAE latent reconstruction $a_{\mathrm{src}}$, which initializes autoregressive decoding through $\mathcal{P}_{init}$ and conditions the caption embedding through $\mathcal{P}_{cond}$. (b) Zoom-in of CAM inside $\mathcal{P}_{cond}$ in (a): CAM summarizes the retained-scale latent feature with learned queries, lets the text state attend to these summaries, and uses gated fusion to produce the retained-scale-aware condition $c_{\mathrm{align}}$.}
    \label{fig:method_flowchart}
\end{figure}

We propose \textbf{SAVAR}, an autoregressive compression framework for extreme-low bitrate image compression. As shown in Figure~\ref{fig:method_flowchart}, SAVAR transmits early-scale tokens and a compact caption. The received tokens provide image-specific layout information, while the caption supplies global semantics. Instead of feeding the caption to the decoder as a standalone prompt, SAVAR first reconstructs a partial VQ-VAE latent from the retained scales. Concretely, SAVAR uses this retained-scale latent in two paths: one path maps it to the $\mathrm{SOS}$ initialization, and the other fuses the decoded caption embedding with the same retained-scale feature before AdaLN control of the VAR backbone. This separates the autoregressive initial state from the global semantic condition while grounding both in the transmitted visual tokens. The design helps the decoder use the caption for semantic details that are missing from the transmitted tokens while reducing the risk of overriding the layout specified by those tokens.

Our contributions are:
\begin{itemize}
    \item We characterize \emph{semantic drift} as a failure mode in this regime and provide evidence that caption guidance is more effective when anchored to transmitted scale tokens than when used as a standalone prompt.
    \item We introduce SAVAR, which uses retained scale tokens for $\mathrm{SOS}$ initialization and conditions caption features on the partial VQ-VAE latent reconstruction before AdaLN; the bitstream transmits only early-scale tokens and caption text, with no hidden embedding side channel.
    \item On the Kodak and CLIC2020 datasets, SAVAR reduces LPIPS by $8$--$9\%$ and DISTS by up to $22\%$ relative to matched ARPC \cite{zhang2026arpc} retained-scale operating points, while using a lower measured token-plus-caption bpp and decoding $512 \times 512$ images in $218$ ms.
\end{itemize}
\section{Related Work}
\label{sec:related_work}

\noindent\textbf{Learned Image Compression.}
Learned image compression initially focused on the rate-distortion tradeoff, usually measured by MSE, MS-SSIM, or related fidelity metrics. Most early systems used autoencoder architectures with learned entropy models. Ball\'{e} \textit{et al.} \cite{balle2018variationalimagecompressionscale} introduced the scale hyperprior to model spatial dependencies in latent space. Later work improved compression efficiency with joint autoregressive contexts \cite{minnen2018joint}, attention mechanisms \cite{Cheng_2020_CVPR}, real-time neural codecs \cite{wang2023evc}, and multi-reference entropy modeling \cite{jiang2023mlic}. These models have reported strong objective-rate results relative to traditional codecs such as EVC and VVC \cite{choi2020overview, bross2021overview}, but distortion-optimized training often removes high-frequency textures in the extreme-low bitrate setting.

\noindent\textbf{Generative and Perceptual Compression.}
Generative compression addresses this texture loss by using image priors, following the rate-distortion-perception tradeoff \cite{blau2019rethinking}. Early methods used Generative Adversarial Networks \cite{goodfellow2014generative, Agustsson_2019_ICCV, mentzer2020high} to synthesize realistic textures. MS-ILLM targets statistical fidelity with an implicit local likelihood discriminator \cite{muckley2023improving}. Diffusion-based codecs subsequently incorporated latent diffusion priors \cite{rombach2021highresolution} into compression, including conditional diffusion compression \cite{yang2024lossy}, PerCo \cite{careil2023towards}, DiffEIC \cite{li2024extremeimagecompressionlatent}, and RDEIC \cite{li2024diffusionbasedextremeimagecompression}. These methods can synthesize realistic details in the extreme-low bitrate setting, but iterative denoising increases decoding latency. When the transmitted representation is severely quantized, diffusion decoders may also deviate from the source image and synthesize plausible but incorrect content.

\noindent\textbf{Autoregressive Modeling and VAR.}
AR models over visual tokens provide another route to generative compression. VQ-VAE \cite{van2017neural} and VQGAN \cite{esser2021taming} showed that images can be represented by discrete codebook indices. MaskGIT \cite{chang2022maskgit} and compact tokenizers such as TiTok \cite{yu2024imageworth32tokens} further showed that these tokens can support efficient high-resolution synthesis. Tokenizer-based compression builds on the same observation. Fine-tuned VQGAN codecs encode images as codebook indices \cite{mao2024extreme}. GLC performs transform coding in a generative VQ-VAE latent space \cite{Jia_2024_CVPR}. UIGC uses a multi-stage transformer to model and recover VQ token maps \cite{xue2024unifying}. HybridFlow combines a codebook stream with continuous features for fidelity preservation \cite{lu2024hybridflow}, and DLF decomposes generative latents into semantic and detail branches in the sub-$0.01$-bpp range \cite{xue2025dlf}. VAR modeling \cite{xue2024visual} reformulates next-token prediction as next-scale prediction for faster generation across spatial resolutions.

Original VAR architectures are designed for generation, not progressive decoding under a bitrate constraint. ARPC \cite{zhang2026arpc} adapts VAR to image compression by truncating the discrete token stream at different scales. This makes autoregressive compression a low-latency alternative to iterative diffusion codecs. A limitation appears in the extreme-low bitrate regime: in ARPC, the same discrete tokens carry both global semantics and local structure. When only the first few coarse scales are transmitted, the received tokens provide a rough description of the scene. The autoregressive model then synthesizes missing semantics from its prior, which can result in source-inconsistent objects, layouts, or textures.

\noindent\textbf{Semantic-Guided Compression.}
Semantic guidance has been explored in generative compression, where text descriptions or semantic features condition reconstruction \cite{lee2024neuralimagecompressiontextguided, careil2023towards}. Key design choices include whether the semantic signal is transmitted, how its rate is counted, and whether the decoder conditions on a global semantic prior or on information aligned with the received image representation. In our setting, caption information is not free side information: the transmitted caption bytes are included in the total bpp. The Jina-CLIP-v2 embedding is computed at the decoder after the caption text is decoded, so no dense text embedding is transmitted.
\section{Methodology}
\label{sec:methodology}

\subsection{Problem Setup and Bitstream}

\textbf{Task Formulation.}
SAVAR operates on the multi-scale discrete token representation used by VAR modeling \cite{xue2024visual}. Given an input image $I$, a frozen tokenizer represents it as token maps $\{r_1, r_2, \dots, r_K\}$ across $K$ progressively larger spatial scales.

Each token map $r_i \in \{1, \dots, V\}^{h_i \times w_i}$ has a spatial size of $h_i \times w_i$ and a vocabulary size of $V$. Given the retained scales, the decoder autoregressively completes the missing part of the token pyramid:
\begin{equation}
    P(r_{m+1:K} \mid r_{\le m}) =
    \prod_{i=m+1}^{K} P(r_i \mid r_{<i}).
\end{equation}
At the target extreme-low bitrate setting, SAVAR transmits only the first $m$ scales, denoted as $r_{\le m}$, alongside a compact caption $T$ as textual side information. The decoder reconstructs the missing high-resolution scales $\{r_{m+1}, \dots, r_K\}$ from the received bitstream and then decodes the completed token pyramid into the output image. Each operating point is determined by the number of retained scales, which dictates the token rate.

\textbf{Bitstream and Caption Side Information.}
The caption side information is transmitted as plain text, not as an embedding. A frozen BLIP-2 captioner \cite{pmlr-v202-li23q} generates a short caption $T$ from the input image at the encoder, and this text is sent alongside the coded scale tokens. The decoder recovers the text and applies a frozen Jina-CLIP-v2 text encoder \cite{gunther2024jina} to obtain $e_{\mathrm{text}}$. The captioner and the input image are not available at the decoder.

The transmitted bitstream is
\begin{equation}
    B = B_{\mathrm{token}} + B_{\mathrm{text}},
\end{equation}
where $B_{\mathrm{token}}$ contains the retained VAR scale-token maps and $B_{\mathrm{text}}$ contains the caption bytes. In the reported experiments, each VAR index is encoded using a fixed 12-bit code because the VQ vocabulary has $V=4096$ entries, i.e.,
\begin{equation}
    B_{\mathrm{token}} = \sum_{i=1}^{m} h_i w_i \lceil \log_2 V \rceil .
\end{equation}
This fixed-length code defines the token rate in our implementation. With an adaptive entropy coder, $B_{\mathrm{token}}$ is replaced by the realized entropy-coded length. The text payload is counted as
\begin{equation}
    B_{\mathrm{text}} = 8\, |\mathrm{UTF8}(T)| ,
\end{equation}
without any uncounted prompt or embedding side channels. The resulting total bitrate is
\begin{equation}
    R_{\mathrm{bpp}} =
    \frac{B_{\mathrm{token}} + B_{\mathrm{text}}}{H W},
\end{equation}
where $H \times W$ is the image resolution. The transmitted bitstream does not contain a dense 1024-dimensional Jina-CLIP embedding. If the captioner returns an empty or invalid caption, the encoder sends an empty string; this fallback is evaluated in Appendix~\ref{app:text_condition_robustness}.

\subsection{SAVAR}

The SAVAR pipeline is summarized in Figure~\ref{fig:method_flowchart}(a). Starting from the received scale tokens, the decoder recovers a partial VQ-VAE latent reconstruction $a_{\mathrm{src}}$. SAVAR uses this retained-scale latent feature in two paths: \textbf{$\mathbf{SOS}$ initialization from retained tokens} ($\mathcal{P}_{init}$) and \textbf{retained-scale-aware text conditioning} ($\mathcal{P}_{cond}$).

The first path maps $a_{\mathrm{src}}$ to the $\mathrm{SOS}$ token that initializes autoregressive decoding. The second path fuses the decoded caption embedding with $a_{\mathrm{src}}$ before applying AdaLN conditioning in the VAR backbone. The resulting completion distribution is:
\begin{equation}
    P(r_{m+1:K} \mid r_{\le m}, c_{\mathrm{align}})
    = \prod_{i=m+1}^K P(r_i \mid r_{<i}, c_{\mathrm{align}}),
\end{equation}
where $c_{\mathrm{align}}$ is a retained-scale-aware text condition rather than a text-only prompt embedding.

\textbf{$\mathbf{SOS}$ Initialization from Retained Tokens ($\mathcal{P}_{init}$).}
SAVAR computes the $\mathrm{SOS}$ token from the continuous latent features reconstructed from the received scale tokens.

Let $a_{\mathrm{src}} \in \mathbb{R}^{C_v \times H_v \times W_v}$ denote the partial VQ-VAE latent reconstruction obtained from the transmitted scale tokens. For $512 \times 512$ images, the VQ-VAE latent grid is $H_v=W_v=32$ with $C_v=32$ channels. Given the transmitted token maps $r_{\le m}$, we map each token to its VQ embedding, upsample each scale to a $32 \times 32$ resolution, pass it through the corresponding residual quantization convolution, and sum the scale contributions in the VQ-VAE latent space. Thus, $a_{\mathrm{src}}$ is reconstructed before the post-quantization convolution and image decoder. It is not a decoded RGB image, but it carries coarse layout and appearance information derived from the input image.

We define the $\mathrm{SOS}$ initialization as:
\begin{equation}
    \mathrm{SOS} = \phi_{\mathrm{init}}(a_{\mathrm{src}}),
\end{equation}
where $\phi_{\mathrm{init}}$ uses two $3 \times 3$ convolutional layers that map 32 channels to 64 and then 96 channels. The output is flattened, projected to 768 dimensions, normalized, and projected to the VAR hidden dimension. This vector initializes autoregressive decoding before any high-resolution token is generated.

\begin{figure}[!t]
    \centering
    \includegraphics[width=\linewidth]{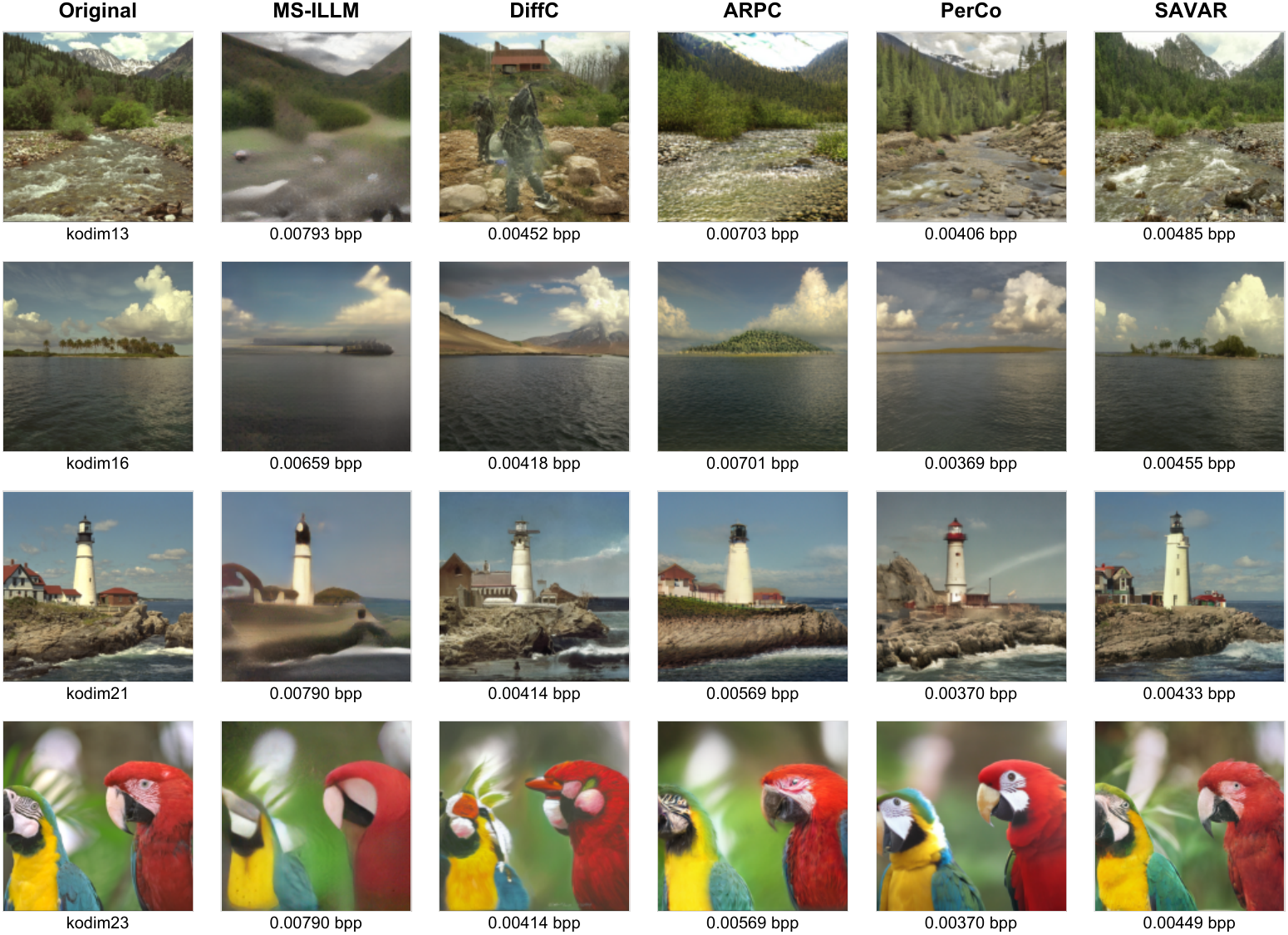}
    \caption{Qualitative comparison on Kodak at the five-scale SAVAR operating point. With few retained scale tokens, some reconstructions keep coarse layout while changing object identity or texture. In these examples, SAVAR shows fewer such substitutions when the caption condition is anchored to retained-scale latent features.}
    \label{fig:visual_comparison}
\end{figure}

Before detailing the conditioning path, a qualitative preview of the source-inconsistent substitutions that can appear at this operating point and how retained-scale-aware caption conditioning changes these examples is given in Figure~\ref{fig:visual_comparison}.

\subsection{Retained-Scale-Aware Conditioning}

As detailed in Figure~\ref{fig:method_flowchart}(b), the conditioning path $\mathcal{P}_{cond}$ is implemented by CAM, which maps the decoded caption and the retained-scale latent feature to a retained-scale-aware text condition, $c_{\mathrm{align}}$.

On the decoder side, we process the decoded caption $T$ through a frozen Jina-CLIP-v2 \cite{gunther2024jina} text encoder to obtain a 1024-dimensional embedding $e_{\mathrm{text}}$. The conditioning path uses the same retained-scale latent feature $a_{\mathrm{src}} \in \mathbb{R}^{32 \times H_v \times W_v}$ reconstructed from the retained VQ-VAE scale tokens. The CAM maps both signals into a 1024-dimensional condition:
\begin{equation}
    c_{\mathrm{align}} = \mathrm{CAM}(e_{\mathrm{text}}, a_{\mathrm{src}}).
\end{equation}
CAM first flattens $a_{\mathrm{src}}$ into spatial tokens and projects each 32-dimensional token to a 512-dimensional hidden space. The 1024-dimensional text embedding is projected to the same space. Eight learnable summary queries attend to these spatial tokens with multi-head attention, producing image-conditioned summary tokens. A single text token then attends to these summary tokens to obtain a retained-scale spatial update. CAM predicts a sigmoid gate from the text state and the spatial update, applies the gated update to the text token, and fuses the result with LayerNorm and a feed-forward network. A final projection maps the fused hidden state back to 1024 dimensions, yielding $c_{\mathrm{align}}$.

This retained-scale-aware condition is then linearly projected to the VAR hidden dimension and modulates the activations of every Transformer block through AdaLN:
\begin{equation}
    \text{AdaLN}(h, c_{\mathrm{align}}) = \gamma(c_{\mathrm{align}}) \odot \text{LayerNorm}(h) + \beta(c_{\mathrm{align}})
\end{equation}
where $h$ is the hidden state of the Transformer block, and $\gamma$ and $\beta$ are learned affine transformations. 

\textbf{CAM Training Objective.}
We train CAM as a standalone alignment module with the VQ-VAE and Jina-CLIP-v2 frozen. Given $a_{\mathrm{src}}$ and $e_{\mathrm{text}}$, CAM predicts $c_{\mathrm{align}}$ and is supervised by the normalized Jina-CLIP-v2 image embedding $e_{\mathrm{img}}$:
\begin{equation}
    \mathcal{L}_{\mathrm{CAM}}
    =
    \mathcal{L}_{\mathrm{pair}}
    +
    w(t)\mathcal{L}_{\mathrm{ctr}},
\end{equation}
where the pairwise term is
\begin{equation}
    \mathcal{L}_{\mathrm{pair}}
    =
    \frac{1}{N}\sum_{n=1}^{N}
    \left(1 - \left\langle
    \bar{c}_{\mathrm{align}}^{(n)}, \bar{e}_{\mathrm{img}}^{(n)}
    \right\rangle\right),
\end{equation}
bars denote $\ell_2$ normalization, and $\mathcal{L}_{\mathrm{ctr}}$ is a symmetric CLIP-style batch contrastive loss between CAM outputs and frozen image embeddings. Additional architecture, training, and decoding details are given in Appendix~\ref{app:reproducibility_details}.
\section{Experiments}
\label{sec:experiments}

\subsection{Experimental Setup}

We evaluate SAVAR below $0.01$ bpp under a fixed preprocessing and metric protocol.

\textbf{Datasets.} We evaluate SAVAR on the Kodak \cite{kodak}, CLIC2020 validation \cite{CLIC2020}, and MSCOCO30k \cite{lin2014microsoft} datasets. All three datasets use the same image preprocessing: the shorter side is resized to 512 pixels with LANCZOS interpolation, followed by a $512 \times 512$ center crop.

\textbf{Implementation Details.} 
The encoder uses frozen BLIP-2 \cite{pmlr-v202-li23q} to generate a caption and transmits it with the retained VAR scale tokens. The decoder computes the Jina-CLIP-v2 text embedding from the received caption; no dense embedding is transmitted.

The autoregressive Transformer backbone is initialized from pretrained VAR-d36 weights and fine-tuned for reconstruction from retained scales. Training uses ImageNet ILSVRC train images paired with BLIP-2-generated captions for 15 epochs, ImageNet ILSVRC val images and their generated captions for validation, AdamW with a linear LR schedule, condition dropout 0.0, and random seed 42. Optimizer groups, training-step count, parameter counts, decoding protocol, and BLIP-2 captioning settings are given in Appendix~\ref{app:reproducibility_details}. CAM pretraining uses the loss in Section~\ref{sec:methodology}, with the contrastive weight annealed from $0.04$ to $0.004$ over the first six epochs.

Token bpp counts 12 bits per VAR token, and text bpp counts 8 bits per caption byte. We evaluate two operating points: four retained scales with 30 transmitted tokens and $0.001373$ token bpp, and five retained scales with 66 transmitted tokens and $0.003021$ token bpp.

\textbf{Bitrate Accounting.}
For SAVAR, the total bpp is the sum of the fixed-length VAR token rate and the measured caption rate. The decoder receives only the caption and retained scale tokens, and captions are generated by the encoder from the test image rather than taken from dataset labels. SAVAR reconstructions use the fixed decoding configuration specified for each operating point, with no side information. For all baselines, we report end-to-end bpp at the corresponding operating point.

ARPC uses a structurally modified VQ-VAE, whereas SAVAR uses the native VAR VQ-VAE. Because these tokenizers have different early-scale patch schedules and entropy-coding mechanisms, retained-scale matching should be interpreted as a structural operating-point comparison rather than a equal-tokenizer or equal-rate comparison. Text-condition ablations with correct, missing, generic, and shuffled captions are reported in Appendix~\ref{app:ablation_studies}.

\textbf{Baselines.} Our autoregressive baseline is ARPC \cite{zhang2026arpc}, reproduced using the official code and checkpoints at retained-scale counts matched to SAVAR. For extreme-low bitrate quality comparisons, we also compare with PerCo \cite{careil2023towards}, MS-ILLM \cite{muckley2023improving}, DiffC \cite{yang2024lossy}, and DLF \cite{xue2025dlf}. For these baselines, we use official reconstructions when available; otherwise, we run the official inference code or checkpoints. No baseline model is trained by us. The shared preprocessing and metric protocol are detailed in Appendix~\ref{app:evaluation_protocol}.

\textbf{Metrics.} We report PSNR and MS-SSIM for signal fidelity, LPIPS and DISTS for perceptual quality, FID/KID for realism, and CLIP-I for source-image semantic similarity. All reported FID/KID values use the same Inception-v3 pool3 feature extractor and the same real/reconstruction preprocessing. We treat FID/KID as auxiliary metrics, especially on MSCOCO30k; the paired reconstruction metrics are LPIPS, DISTS, MS-SSIM, and CLIP-I. CLIP-I is reported only as a source-image similarity metric under a fixed evaluation protocol. Decoding time is measured on one NVIDIA L20X GPU.

\subsection{Extreme-Low Bitrate Reconstruction Quality}

This experiment evaluates SAVAR as a codec in the extreme-low bitrate regime. We report total bpp for each method, focus the main tables on results below $0.01$ bpp, and reserve higher-bpp operating points only for curve context.

\begin{figure}[htbp]
    \centering
    \includegraphics[width=\linewidth]{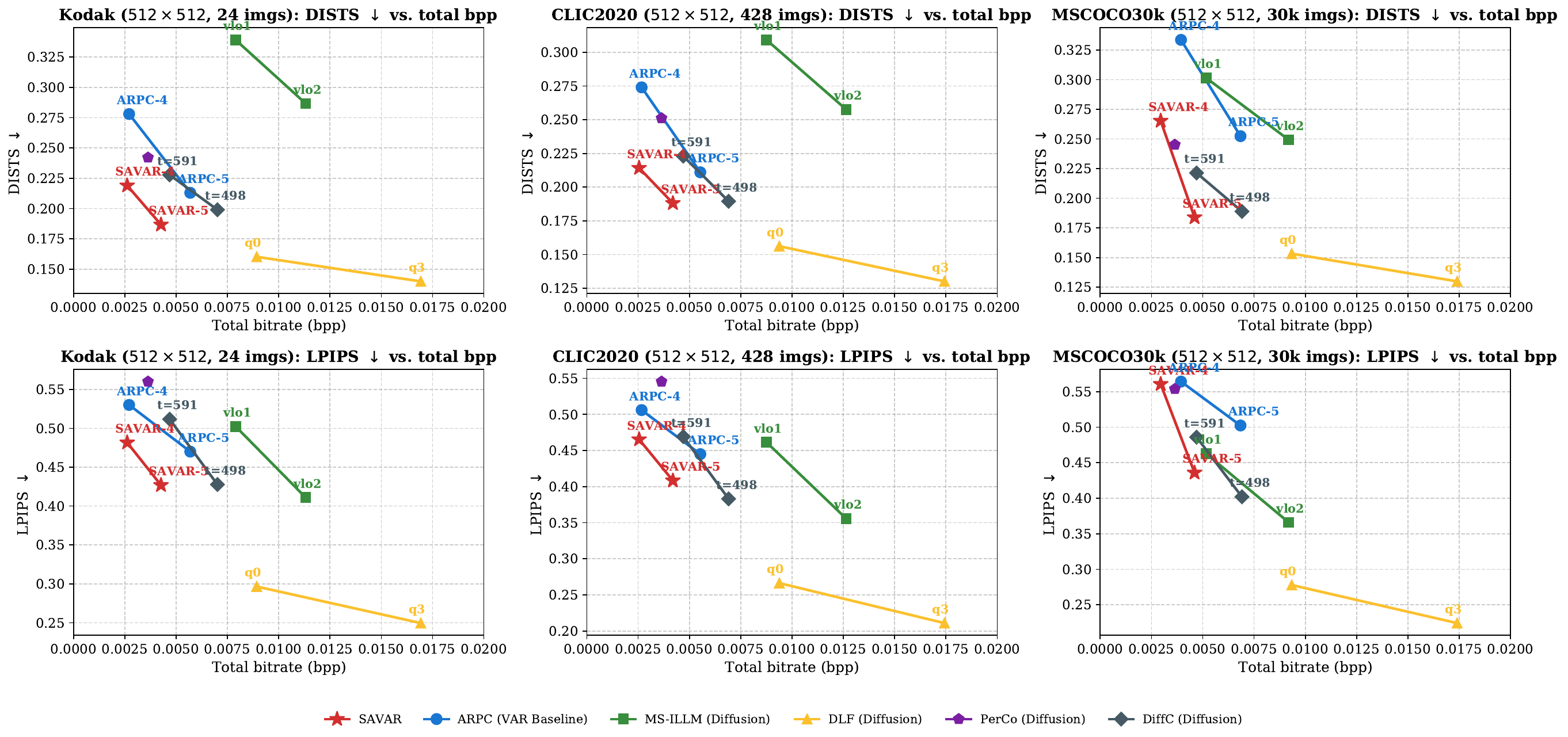}
    \caption{Rate-distortion-perception curves comparing SAVAR with the autoregressive baseline ARPC and generative/perceptual compression baselines at $512 \times 512$ resolution, using total bpp. Columns show Kodak-24, CLIC2020, and MSCOCO30k; rows show DISTS and LPIPS. SAVAR and ARPC are compared at retained-scale operating points; PerCo, MS-ILLM, DiffC, and DLF are shown at their available extreme-low bitrate operating points. MS-ILLM is restricted to its two lowest-bpp points, DLF includes its adjacent point below $0.02$ bpp for curve context, and PerCo is restricted to its lowest-bpp point.}
    \label{fig:rd_curves}
\end{figure}

The DISTS--bpp and LPIPS--bpp curves on Kodak, CLIC2020, and MSCOCO30k are plotted in Figure~\ref{fig:rd_curves}, comparing SAVAR against the autoregressive baseline ARPC and generative/perceptual codecs including PerCo, MS-ILLM, DiffC, and DLF. The ARPC comparison controls for the number of transmitted progressive scales, while the generative/perceptual baselines provide context for the broader sub-$0.01$-bpp compression regime. At the retained-scale operating points most directly matched to SAVAR's design, SAVAR lowers DISTS and LPIPS relative to ARPC while using comparable or fewer bits. DISTS captures perceptual structure and texture similarity, while LPIPS measures learned perceptual distance; improvements on both metrics are consistent with source-aligned perceptual reconstruction under this evaluation protocol. Complementary CLIP-I, FID, and KID curves are reported in Appendix~\ref{app:auxiliary_curves}.

\textbf{Qualitative Results.}
SAVAR is visually compared with MS-ILLM, DiffC, ARPC, and PerCo in Figure~\ref{fig:visual_comparison} to illustrate the same source-inconsistency issue suggested by the curves, with CLIC2020 and MSCOCO30k examples provided in Figure~\ref{fig:appendix_large_dataset_visual_comparison}. Direct autoregressive reconstruction is ambiguous because the early scale tokens mainly carry coarse layout and limited semantic and appearance cues. In these examples, retained-scale-aware caption conditioning yields fewer source-inconsistent substitutions, consistent with the design goal of using transmitted scale tokens for layout anchoring and captions for missing semantic details.

The values behind the Kodak and CLIC2020 distortion curves are reported in Table~\ref{tab:comprehensive_results}, and the extreme-low bitrate MSCOCO30k results are given in Table~\ref{tab:mscoco_results} in Appendix~\ref{app:mscoco_extreme_low_bitrate}. We compare SAVAR at two operating points with ARPC using the same retained-scale counts (4 and 5 retained scales), and include all baseline operating points in the target bpp range. This retained-scale comparison matches the number of transmitted progressive scales, but it does not by itself attribute bitrate differences to the SAVAR conditioning mechanism, because ARPC and SAVAR use different VQ-VAE/tokenizer designs.

Under this retained-scale comparison, SAVAR lowers the matched ARPC points on the perceptual metrics emphasized here. On CLIC2020, SAVAR lowers LPIPS/DISTS from $0.506/0.274$ to $0.465/0.214$ at 4 retained scales and from $0.445/0.211$ to $0.408/0.188$ at 5 retained scales, while using comparable or fewer bits. The same pattern appears on Kodak, where SAVAR reduces DISTS from $0.278$ to $0.219$ at 4 retained scales and from $0.213$ to $0.187$ at 5 retained scales. On MSCOCO30k, SAVAR reduces DISTS from $0.3336$ to $0.2653$ at 4 retained scales and from $0.2524$ to $0.1839$ at 5 retained scales, and raises CLIP-I from $0.5757$ to $0.6370$ and from $0.6581$ to $0.8422$, respectively. Compared with the generative/perceptual baselines in the same extreme-low bitrate range, SAVAR also reaches lower DISTS than DiffC on CLIC2020 at both SAVAR operating points and on MSCOCO30k at the 5-retained-scale point.

\begin{table}[htbp]
    \caption{Quantitative comparison on Kodak and CLIC2020 at $512 \times 512$ in the target extreme-low bitrate range. BPP denotes the token-plus-caption bitrate for SAVAR and the end-to-end bitrate for each baseline operating point. \textbf{Bold} marks the best value within each matched ARPC/SAVAR operating-point pair.}
    \label{tab:comprehensive_results}
    \centering
    \resizebox{\textwidth}{!}{
    \begin{tabular}{c}
    \begin{tabular}{ll l c cccccc}
        \multicolumn{10}{c}{\small\textbf{Kodak (24 imgs)}} \\
        \toprule
        Architecture & Method & Operating point & BPP $\downarrow$ & PSNR $\uparrow$ & MS-SSIM $\uparrow$ & LPIPS $\downarrow$ & DISTS $\downarrow$ & FID $\downarrow$ & KID $\downarrow$ \\
        \midrule
        \multirow{4}{*}{\textit{Autoregressive}} 
        & ARPC \cite{zhang2026arpc} & 4 retained scales & 0.002710 & 14.99 & 0.376 & 0.530 & 0.278 & -- & -- \\
        & \textbf{SAVAR} & 4 retained scales & \textbf{0.002615} & \textbf{15.87} & \textbf{0.381} & \textbf{0.482} & \textbf{0.219} & -- & -- \\
        \cmidrule{2-10}
        & ARPC \cite{zhang2026arpc} & 5 retained scales & 0.005690 & 16.47 & \textbf{0.449} & 0.470 & 0.213 & -- & -- \\
        & \textbf{SAVAR} & 5 retained scales & \textbf{0.004263} & \textbf{16.52} & 0.419 & \textbf{0.427} & \textbf{0.187} & -- & -- \\
        \midrule
        \multirow{5}{*}{\textit{Generative/perceptual}} 
        & PerCo \cite{careil2023towards} & $low$ & 0.003630 & 15.14 & 0.377 & 0.560 & 0.242 & -- & -- \\
        & MS-ILLM \cite{muckley2023improving} & $vlo1$ & 0.007899 & 20.23 & 0.593 & 0.502 & 0.339 & 245.01 & 0.0426 \\
        & DLF \cite{xue2025dlf} & $q0$ & 0.008928 & 18.65 & 0.587 & 0.297 & 0.160 & 110.78 & -- \\
        & DiffC \cite{yang2024lossy} & $t591$ & 0.004684 & 16.91 & 0.451 & 0.512 & 0.228 & -- & -- \\
        & DiffC \cite{yang2024lossy} & $t498$ & 0.007014 & 18.31 & 0.545 & 0.428 & 0.199 & -- & -- \\
        \bottomrule
    \end{tabular} \\
    \vspace{1em} \\
    \begin{tabular}{ll l c cccccc}
        \multicolumn{10}{c}{\small\textbf{CLIC2020 (428 imgs)}} \\
        \toprule
        Architecture & Method & Operating point & BPP $\downarrow$ & PSNR $\uparrow$ & MS-SSIM $\uparrow$ & LPIPS $\downarrow$ & DISTS $\downarrow$ & FID $\downarrow$ & KID $\downarrow$ \\
        \midrule
        \multirow{4}{*}{\textit{Autoregressive}} 
        & ARPC \cite{zhang2026arpc} & 4 retained scales & 0.002670 & 13.94 & 0.374 & 0.506 & 0.274 & 116.44 & 0.0062 \\
        & \textbf{SAVAR} & 4 retained scales & \textbf{0.002547} & \textbf{14.82} & \textbf{0.396} & \textbf{0.465} & \textbf{0.214} & \textbf{97.10} & \textbf{0.0017} \\
        \cmidrule{2-10}
        & ARPC \cite{zhang2026arpc} & 5 retained scales & 0.005530 & 15.55 & \textbf{0.458} & 0.445 & 0.211 & 102.52 & 0.0042 \\
        & \textbf{SAVAR} & 5 retained scales & \textbf{0.004194} & \textbf{15.59} & 0.443 & \textbf{0.408} & \textbf{0.188} & \textbf{90.87} & \textbf{0.0007} \\
        \midrule
        \multirow{5}{*}{\textit{Generative/perceptual}} 
        & PerCo \cite{careil2023towards} & $low$ & 0.003640 & 14.14 & 0.402 & 0.546 & 0.251 & 103.86 & 0.0011 \\
        & MS-ILLM \cite{muckley2023improving} & $vlo1$ & 0.008757 & 20.38 & 0.654 & 0.461 & 0.309 & 97.73 & 0.0356 \\
        & DLF \cite{xue2025dlf} & $q0$ & 0.009380 & 18.40 & 0.644 & 0.266 & 0.156 & 32.20 & 0.0024 \\
        & DiffC \cite{yang2024lossy} & $t591$ & 0.004700 & 16.80 & 0.519 & 0.469 & 0.223 & 108.57 & 0.0028 \\
        & DiffC \cite{yang2024lossy} & $t498$ & 0.006910 & 18.11 & 0.607 & 0.383 & 0.190 & 89.98 & 0.0005 \\
        \bottomrule
    \end{tabular}
    \end{tabular}
    }
    \vspace{0.5em}
    \raggedright \small \textit{Note:} Baseline operating points denote official checkpoint or timestep names.
\end{table}

\subsection{Mechanism Analysis and Efficiency}

This subsection compares SAVAR against its ablations to check whether SAVAR's gains come from the token-grounded semantic conditioning, then reports the runtime cost of that design.

\textbf{Ablation Summary.}
Replacing token-based $\mathrm{SOS}$ initialization with a text-derived initialization causes a larger degradation than removing CAM alone, while replacing the CAM condition with text also weakens reconstruction. These results support the design choice of anchoring both the autoregressive start state and the global semantic condition to the transmitted tokens (Table~\ref{tab:diagnosis}). In Appendix~\ref{app:text_condition_robustness}, the caption itself is further perturbed; the results show that correct captions provide information mainly when CAM conditions them on the retained-scale latent feature.

\begin{table}[htbp]
    \caption{Path-level ablation on Kodak with the five-scale operating point. All rows use the correct caption and the same total bitrate of $0.004263$ bpp. The full model uses $\mathrm{SOS}$ initialization from retained tokens and retained-scale-aware AdaLN conditioning.}
    \label{tab:diagnosis}
    \centering
    \resizebox{\textwidth}{!}{
    \begin{tabular}{l c c c c c c}
        \toprule
        Variant & $\mathrm{SOS}$ source & AdaLN source & PSNR $\uparrow$ & LPIPS $\downarrow$ & MS-SSIM $\uparrow$ & $\Delta$LPIPS vs. full \\
        \midrule
        \textbf{SAVAR} & tokens & retained-scale & \textbf{16.52} & \textbf{0.4269} & \textbf{0.4191} & 0.0000 \\
        w/o Token-Based Initialization & text & retained-scale & 11.9847 & 0.6964 & 0.1451 & +0.2695 \\
        w/o Retained-Scale AdaLN & tokens & text & 15.7596 & 0.5006 & 0.3592 & +0.0737 \\
        w/o Both & text & text & 11.4489 & 0.7164 & 0.1393 & +0.2895 \\
        \bottomrule
    \end{tabular}
    }
\end{table}

\textbf{Encoding and Decoding Efficiency.} Runtime and a DISTS BD-Rate summary using MS-ILLM as the anchor are reported in Table~\ref{tab:latency}. The BD-Rate is computed from each method's rate-distortion curve, so it complements the retained-scale ARPC/SAVAR comparison with a rate-interpolated summary rather than serving as the primary claim. The rate points, adjacent context points above $0.01$ bpp, and interpolation procedure are specified in Appendix~\ref{app:bdrate_protocol}.

In this summary, SAVAR has the lowest BD-Rate value in Table~\ref{tab:latency} ($-91.41\%$) and decodes a $512 \times 512$ image in $0.218$ s. On the same NVIDIA L20X GPU, its encoder-side runtime decomposes into $0.034$ s for token encoding and $0.112$ s for BLIP-2 caption generation, for a total of $0.146$ s per image before transmission. Since caption generation is performed only at the encoder, the runtime comparison mainly affects decoding, where SAVAR is faster than ARPC and the listed diffusion codecs except MS-ILLM under the reported implementations.

\begin{table}[htbp]
    \caption{Encoding and decoding time on $512 \times 512$ images, with BD-Rate computed on DISTS from rate-distortion curves using MS-ILLM as the anchor. The BD-Rate protocol is given in Appendix~\ref{app:bdrate_protocol}. SAVAR times are measured on one NVIDIA L20X GPU. Lower is better.}
    \label{tab:latency}
    \centering
    \small
    \begin{tabular}{l c c c}
        \toprule
        Method & Total Enc. (s) $\downarrow$ & Dec. Time (s) $\downarrow$ & BD-Rate $\downarrow$ \\
        \midrule
        MS-ILLM \cite{muckley2023improving} & $0.064 \pm 0.010$ & $0.070 \pm 0.011$ & $0.00\%$ \\
        PerCo \cite{careil2023towards} & $0.461 \pm 0.017$ & $2.443 \pm 0.011$ & $-75.89\%$ \\
        DLF \cite{xue2025dlf} & $0.178 \pm 0.015$ & $0.252 \pm 0.014$ & $-69.97\%$ \\
        DiffC \cite{yang2024lossy} & $0.6$--$9.4$ & $2.9$--$8.5$ & $-72.80\%$ \\
        ARPC \cite{zhang2026arpc} & $0.28$--$1.39$ & $1.33$ & $-74.90\%$ \\
        \textbf{SAVAR} & $0.146$ & $0.218 \pm 0.0030$ & $\mathbf{-91.41\%}$ \\
        \bottomrule
    \end{tabular}
\end{table}

\section{Conclusion}
\label{sec:conclusion}
We presented SAVAR, a generative framework for extreme-low bitrate image compression. Direct autoregressive reconstruction can be source-inconsistent when early scale tokens leave object identity underdetermined. SAVAR addresses this by initializing $\mathrm{SOS}$ from transmitted tokens and conditioning the caption embedding on the same partial VQ-VAE latent before AdaLN. This dual-path design reduces perceptual distortion while retaining fast decoding. Limitations include the fixed $512 \times 512$ center-crop evaluation, fixed-length token coding rather than adaptive entropy coding, and reliance on an encoder-side captioner; as with other generative codecs, SAVAR is not intended for forensic or diagnostic reconstruction.

\FloatBarrier
\bibliographystyle{plain}
\bibliography{refs}

\appendix
\section{Auxiliary Rate Curves}
\label{app:auxiliary_curves}

\begin{figure}[!htbp]
    \centering
    \includegraphics[width=\linewidth]{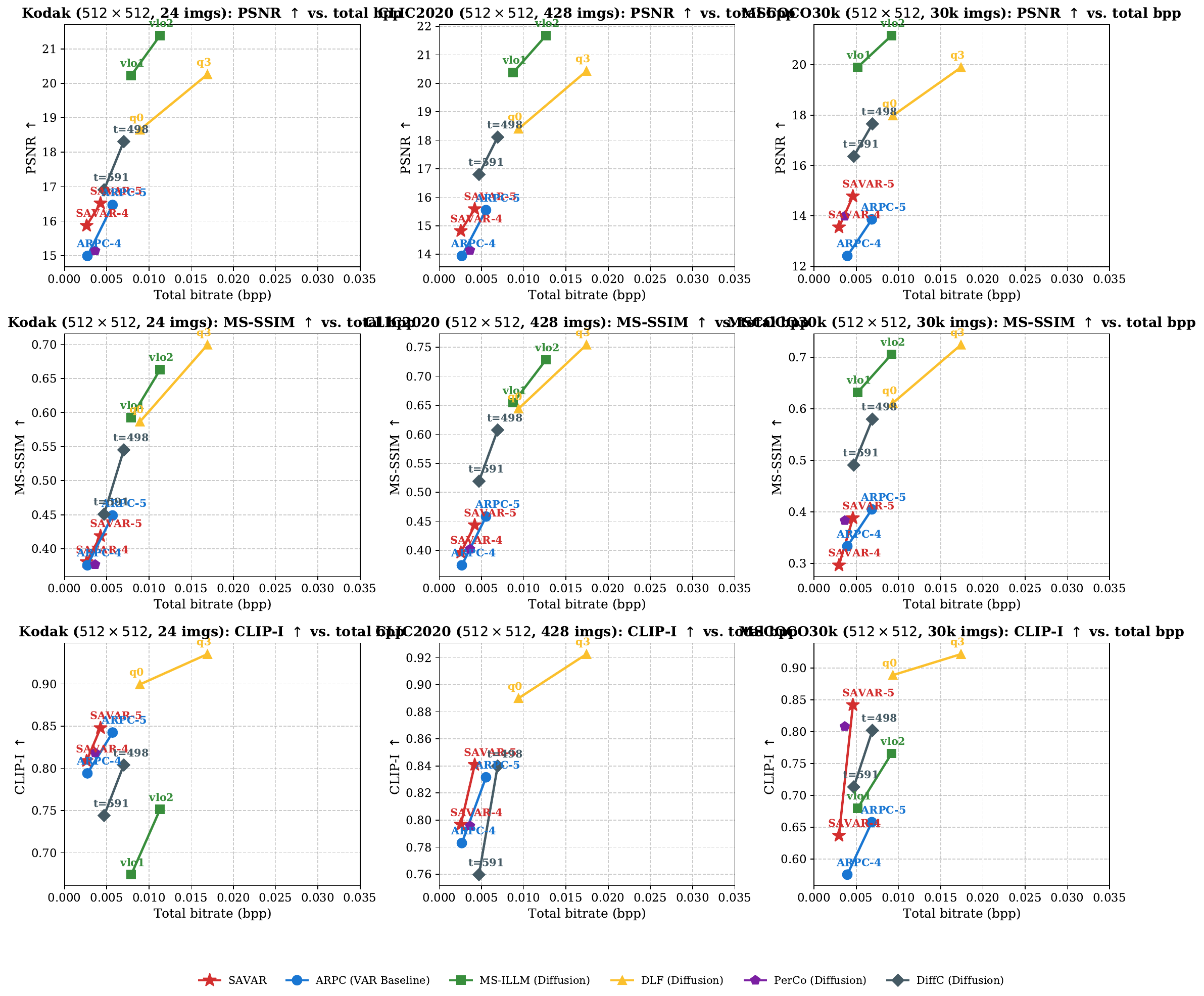}
    \caption{Auxiliary fidelity and CLIP-I rate curves at $512 \times 512$ resolution. Columns show Kodak-24, CLIC2020, and MSCOCO30k; rows show PSNR, MS-SSIM, and CLIP-I.}
    \label{fig:appendix_fidelity_semantic_curves}
\end{figure}

\begin{figure}[!htbp]
    \centering
    \includegraphics[width=\linewidth]{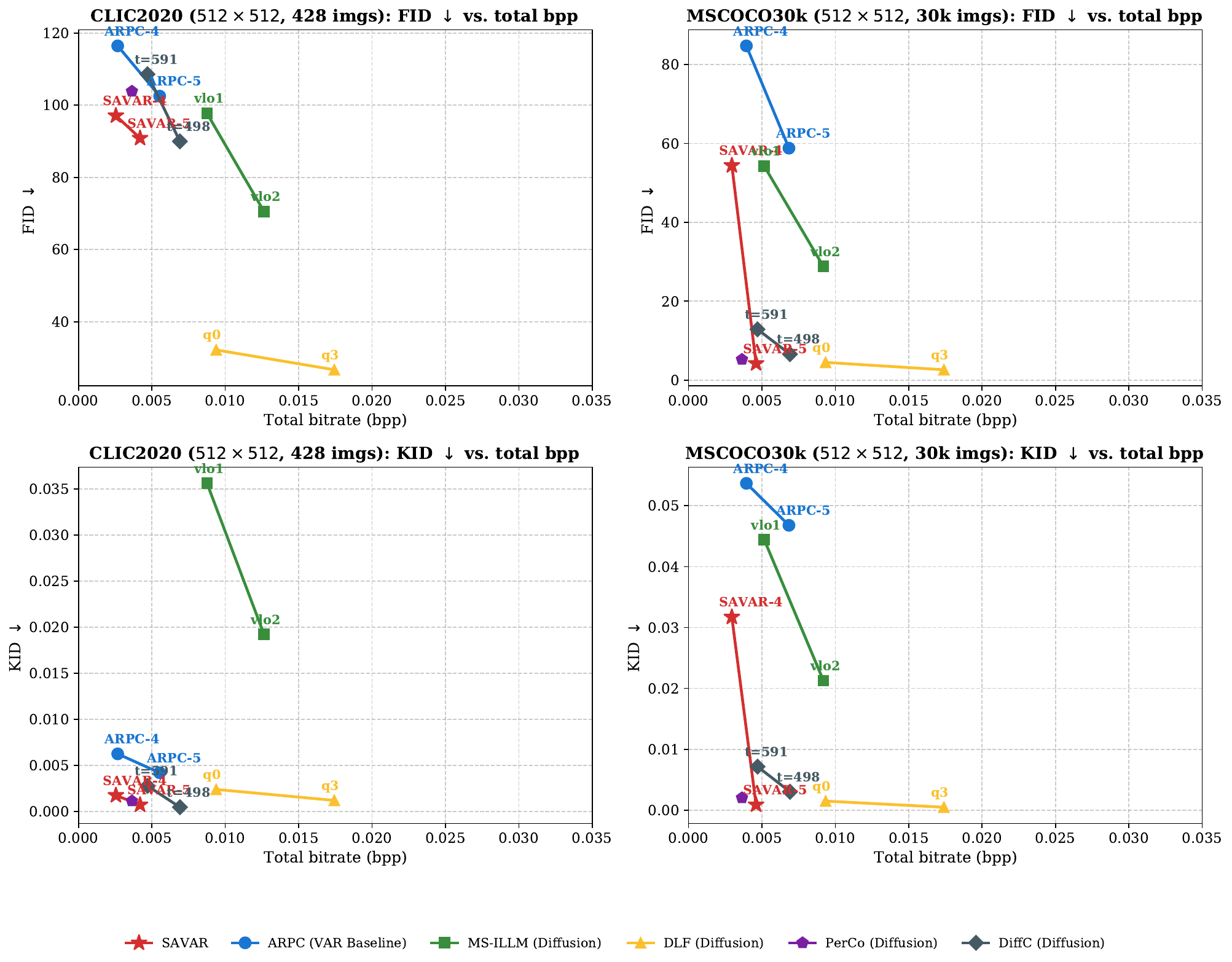}
    \caption{Auxiliary realism rate curves at $512 \times 512$ resolution. Columns show CLIC2020 and MSCOCO30k; rows show FID and KID. Kodak panels are omitted due to limited metric availability.}
    \label{fig:appendix_realism_curves}
\end{figure}

Auxiliary fidelity, CLIP-I, and realism curves that complement the main DISTS/LPIPS comparison in Figure~\ref{fig:rd_curves} are provided in Figures~\ref{fig:appendix_fidelity_semantic_curves} and~\ref{fig:appendix_realism_curves}. The main-text tables retain the extreme-low bitrate reporting protocol defined in Section~\ref{sec:experiments}; the curves include adjacent PerCo and DLF operating points above $0.01$ bpp for visual context, matching the main rate-distortion plots. We omit Kodak FID/KID curves because most methods do not have those aggregate metrics on Kodak.

\section{Ablation Studies}
\label{app:ablation_studies}

\subsection{Text Condition Robustness}
\label{app:text_condition_robustness}

We next examine how much of the full-model gain comes from the caption itself. In Table~\ref{tab:prompt_robustness}, only the transmitted text is perturbed while the retained scale tokens and decoding settings are kept fixed. The no-caption setting transmits no caption and therefore uses the token-only rate of $0.003021$ bpp. The generic-caption and shuffled-caption settings keep the text channel active but remove image-specific or paired semantic content.

\begin{table}[!htbp]
    \caption{Text-condition ablation on Kodak with the five-scale operating point. The full path uses $\mathrm{SOS}$ initialization from retained tokens and retained-scale-aware AdaLN conditioning. The $\mathrm{SOS}$-only path keeps token-based initialization but bypasses CAM. Correct captions improve the full model, while retained scale tokens make the reconstruction less sensitive to weak or mismatched text.}
    \label{tab:prompt_robustness}
    \centering
    \resizebox{\textwidth}{!}{
    \begin{tabular}{l c c c c c c}
        \toprule
        Variant & Caption & BPP $\downarrow$ & PSNR $\uparrow$ & LPIPS $\downarrow$ & MS-SSIM $\uparrow$ & $\Delta$LPIPS vs. full-correct \\
        \midrule
        Full & correct & 0.004263 & \textbf{16.52} & \textbf{0.4269} & \textbf{0.4191} & 0.0000 \\
        Full & no caption & 0.003021 & 15.9850 & 0.4480 & 0.3895 & +0.0211 \\
        Full & generic caption & 0.004263 & 15.9751 & 0.4488 & 0.3898 & +0.0219 \\
        Full & shuffled caption & 0.004263 & 16.0246 & 0.4504 & 0.3901 & +0.0235 \\
        \midrule
        $\mathrm{SOS}$ only & correct & 0.004263 & 15.7596 & 0.5006 & 0.3592 & +0.0737 \\
        $\mathrm{SOS}$ only & no caption & 0.003021 & 15.7575 & 0.4910 & 0.3631 & +0.0641 \\
        $\mathrm{SOS}$ only & generic caption & 0.004263 & 15.7572 & 0.4837 & 0.3611 & +0.0568 \\
        $\mathrm{SOS}$ only & shuffled caption & 0.004263 & 15.6589 & 0.5049 & 0.3510 & +0.0780 \\
        \bottomrule
    \end{tabular}
    }
\end{table}

The caption ablation separates textual side information from the information carried by the retained scale tokens. With the full path, removing or corrupting the caption weakens the reconstruction. Relative to the correct caption, the three perturbed settings lose about $0.50$--$0.54$ dB PSNR, increase LPIPS by $0.0211$--$0.0235$, and reduce MS-SSIM by roughly $0.03$. The generic and shuffled rows keep the text channel active without the paired image description, so their degradation is not explained solely by removing caption bytes. These changes are smaller than the path-level ablation in Table~\ref{tab:diagnosis}, but they indicate that the correct caption contributes information beyond the retained scale tokens in this setting.

The $\mathrm{SOS}$-only rows indicate where this text gain enters the model. When CAM is bypassed, altering the caption has little effect under the same token-based $\mathrm{SOS}$ initialization. Correct, missing, and generic captions all remain near $15.76$ dB PSNR, and shuffled captions cause only a modest additional drop. For every caption mode, the full path has lower LPIPS than the $\mathrm{SOS}$-only path. The largest gap appears under the correct caption, with $0.0737$ lower LPIPS and $0.0599$ higher MS-SSIM. Thus, in this ablation, the text channel is most useful when CAM conditions the caption on the retained-scale latent feature before AdaLN; raw text conditioning alone has a weaker effect.

\section{MSCOCO30k Extreme-Low Bitrate Comparison}
\label{app:mscoco_extreme_low_bitrate}

\begin{table*}[!htbp]
    \caption{Extreme-low bitrate MSCOCO30k comparison at $512 \times 512$.}
    \label{tab:mscoco_results}
    \centering
    \resizebox{\textwidth}{!}{
    \begin{tabular}{ll l c c c c c c c c}
        \toprule
        Architecture & Method & Operating point & BPP $\downarrow$ & PSNR $\uparrow$ & MS-SSIM $\uparrow$ & LPIPS $\downarrow$ & DISTS $\downarrow$ & FID $\downarrow$ & KID $\downarrow$ & CLIP-I $\uparrow$ \\
        \midrule
        \multirow{4}{*}{\textit{Autoregressive}}
        & ARPC \cite{zhang2026arpc} & 4 retained scales & 0.00394 & 12.41 & \textbf{0.3333} & 0.5642 & 0.3336 & 84.73 & 0.05367 & 0.5757 \\
        & \textbf{SAVAR} & 4 retained scales & \textbf{0.00295} & \textbf{13.55} & 0.2963 & \textbf{0.5606} & \textbf{0.2653} & \textbf{54.41} & \textbf{0.03172} & \textbf{0.6370} \\
        \cmidrule{2-11}
        & ARPC \cite{zhang2026arpc} & 5 retained scales & 0.00684 & 13.86 & \textbf{0.4048} & 0.5024 & 0.2524 & 58.80 & 0.04678 & 0.6581 \\
        & \textbf{SAVAR} & 5 retained scales & \textbf{0.00460} & \textbf{14.78} & 0.3880 & \textbf{0.4359} & \textbf{0.1839} & \textbf{4.21} & \textbf{0.00090} & \textbf{0.8422} \\
        \midrule
        \multirow{6}{*}{\textit{Generative/perceptual}}
        & PerCo \cite{careil2023towards} & $low$ & 0.00364 & 13.99 & 0.3832 & 0.5538 & 0.2452 & 5.24 & 0.00203 & 0.8084 \\
        & MS-ILLM \cite{muckley2023improving} & $vlo1$ & 0.00516 & 19.90 & 0.6320 & 0.4630 & 0.3016 & 54.23 & 0.04439 & 0.6800 \\
        & MS-ILLM \cite{muckley2023improving} & $vlo2$ & 0.00919 & 21.16 & 0.7060 & 0.3660 & 0.2493 & 28.80 & 0.02128 & 0.7660 \\
        & DLF \cite{xue2025dlf} & $q0$ & 0.009337 & 17.98 & 0.6113 & 0.2778 & 0.1533 & 4.49 & 0.00150 & 0.8890 \\
        & DiffC \cite{yang2024lossy} & $t591$ & 0.00470 & 16.36 & 0.4908 & 0.4859 & 0.2214 & 12.87 & 0.00716 & 0.7133 \\
        & DiffC \cite{yang2024lossy} & $t498$ & 0.00691 & 17.65 & 0.5799 & 0.4020 & 0.1889 & 6.57 & 0.00303 & 0.8023 \\
        \bottomrule
    \end{tabular}
    }
\end{table*}

\begin{figure}[!t]
    \centering
    \setlength{\tabcolsep}{0pt}
    \renewcommand{\arraystretch}{1.5}
    \begin{tabular}{c}
        \textbf{CLIC2020} \\
        \includegraphics[width=0.76\linewidth]{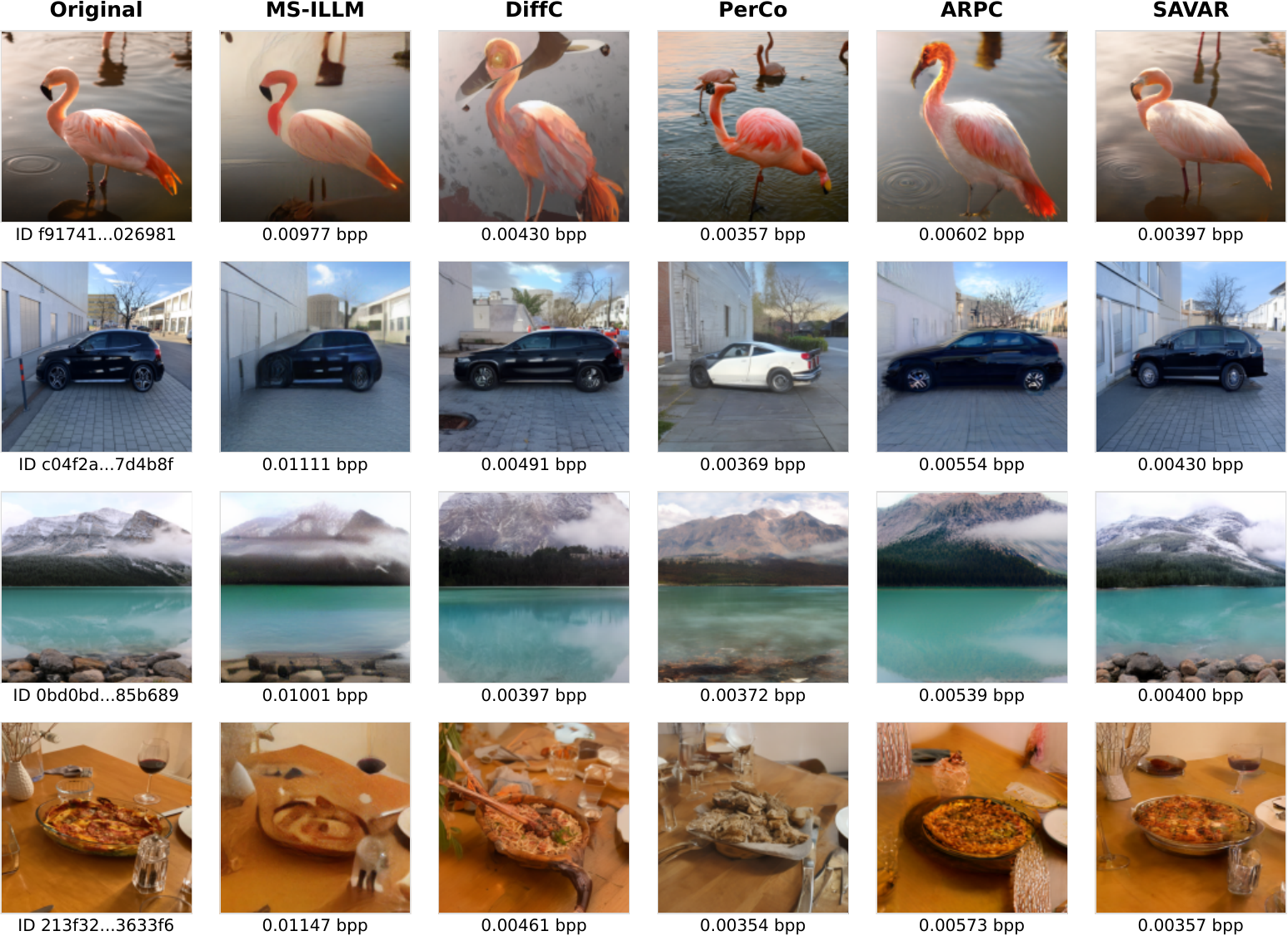} \\
        \midrule
        \textbf{MSCOCO30k} \\
        \includegraphics[width=0.76\linewidth]{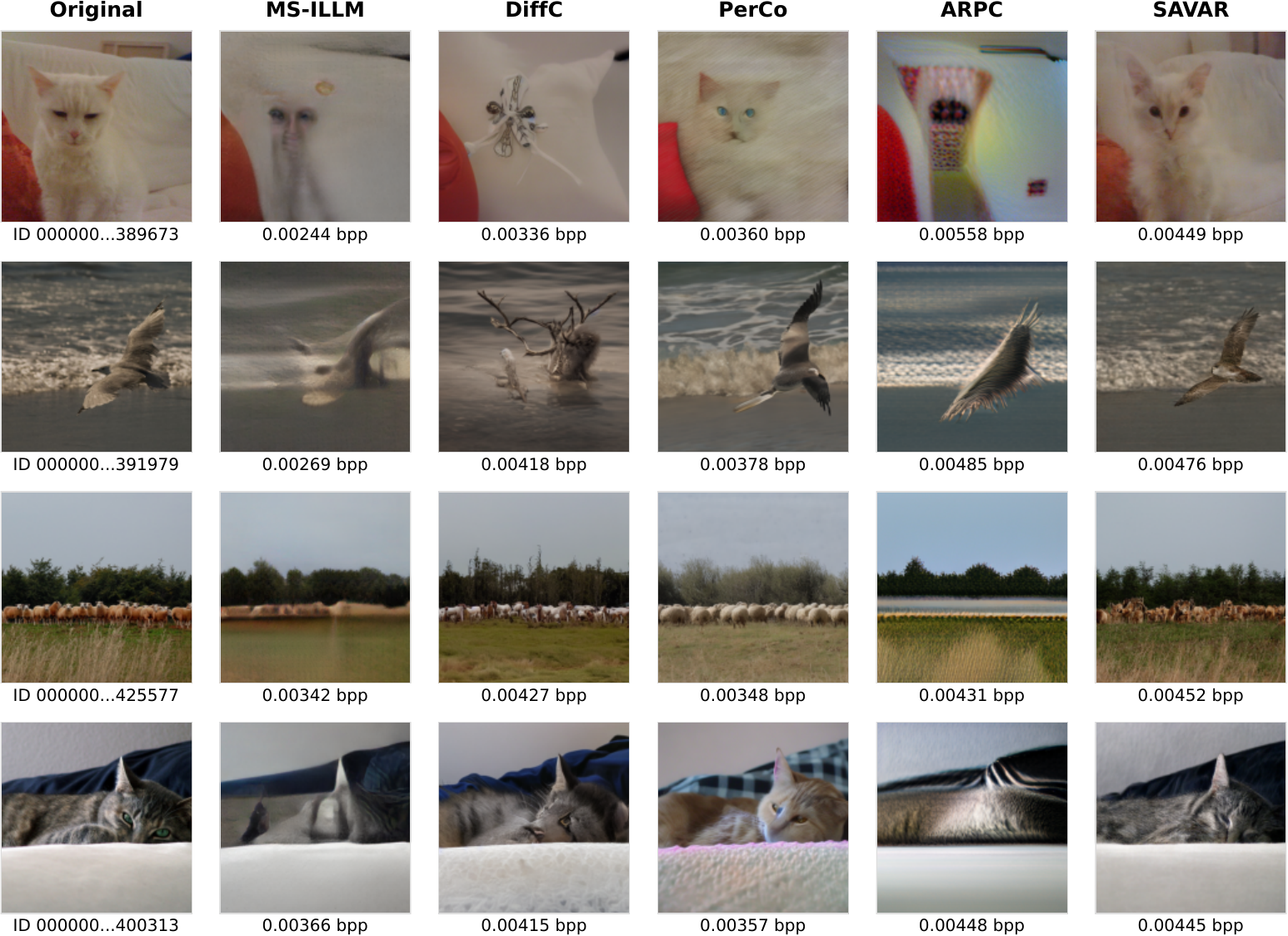}
    \end{tabular}
    \caption{Additional qualitative comparisons on CLIC2020 (top) and MSCOCO30k (bottom) at $512 \times 512$ resolution. Each block shows the source image followed by reconstructions from MS-ILLM, DiffC, PerCo, ARPC, and SAVAR; the bpp printed below each reconstruction is the corresponding end-to-end bitrate for the baseline or token-plus-caption bitrate for SAVAR. These examples complement Tables~\ref{tab:comprehensive_results} and~\ref{tab:mscoco_results} by showing representative source-consistency differences in the sub-$0.01$-bpp regime.}
    \label{fig:appendix_large_dataset_visual_comparison}
\end{figure}

\section{Supplemental CLIP-I and Realism Metrics}
\label{app:complete_semantic_metrics}

CLIP-I and realism metrics for the main extreme-low bitrate operating points and adjacent higher-bpp baseline points used for curve context are reported in Table~\ref{tab:appendix_semantic_realism}. These values complement Tables~\ref{tab:comprehensive_results} and~\ref{tab:mscoco_results}, which emphasize distortion and perceptual fidelity. CLIP-I measures image-to-image semantic similarity between the reconstruction and the source image and is used for within-dataset comparison. Kodak FID/KID entries are left blank in this table because aggregate availability is incomplete across methods and the 24-image set is too small for a stable distribution-level comparison. MS-ILLM is omitted from this table because the complete CLIP-I entries are not available under the same protocol across all three datasets; its distortion and available realism metrics are reported in the main tables.

\begin{table*}[p]
    \caption{Supplemental CLIP-I and realism metrics at $512 \times 512$, including adjacent above-$0.01$ bpp baseline points for curve context. CLIP-I is reported for within-dataset comparison under the fixed evaluation protocol described in Appendix~\ref{app:evaluation_protocol}. Kodak FID/KID entries are omitted here because the metric availability is incomplete across methods.}
    \label{tab:appendix_semantic_realism}
    \centering
    \resizebox{\textwidth}{!}{
    \begin{tabular}{ll l c c c c c}
        \toprule
        Dataset & Method & Operating point & BPP $\downarrow$ & DISTS $\downarrow$ & CLIP-I $\uparrow$ & FID $\downarrow$ & KID $\downarrow$ \\
        \midrule
        \multirow{10}{*}{Kodak-24}
        & \textbf{SAVAR} & 4 retained scales & \textbf{0.002615} & \textbf{0.2190} & 0.8088 & -- & -- \\
        & \textbf{SAVAR} & 5 retained scales & \textbf{0.004263} & \textbf{0.1867} & 0.8478 & -- & -- \\
        & ARPC & 4 retained scales & 0.002710 & 0.2780 & 0.7940 & -- & -- \\
        & ARPC & 5 retained scales & 0.005690 & 0.2130 & 0.8424 & -- & -- \\
        & PerCo & low & 0.003630 & 0.2421 & 0.8183 & -- & -- \\
        & PerCo & mid & 0.032930 & 0.1645 & 0.9100 & -- & -- \\
        & DLF & $q0$ & 0.008928 & 0.1602 & 0.8994 & -- & -- \\
        & DLF & $q3$ & 0.016925 & 0.1399 & 0.9353 & -- & -- \\
        & DiffC & $t591$ & 0.004684 & 0.2278 & 0.7440 & -- & -- \\
        & DiffC & $t498$ & 0.007014 & 0.1990 & 0.8039 & -- & -- \\
        \midrule
        \multirow{10}{*}{CLIC2020}
        & \textbf{SAVAR} & 4 retained scales & \textbf{0.002547} & \textbf{0.2142} & 0.7967 & 97.10 & 0.00174 \\
        & \textbf{SAVAR} & 5 retained scales & \textbf{0.004194} & \textbf{0.1881} & 0.8409 & 90.87 & 0.00071 \\
        & ARPC & 4 retained scales & 0.002670 & 0.2740 & 0.7830 & 116.44 & 0.00624 \\
        & ARPC & 5 retained scales & 0.005530 & 0.2110 & 0.8316 & 102.52 & 0.00419 \\
        & PerCo & low & 0.003640 & 0.2511 & 0.7957 & 103.86 & 0.00113 \\
        & PerCo & mid & 0.032930 & 0.1636 & 0.9121 & 71.31 & -0.00085 \\
        & DLF & $q0$ & 0.009380 & 0.1562 & 0.8900 & 32.20 & 0.00238 \\
        & DLF & $q3$ & 0.017430 & 0.1301 & 0.9226 & 26.70 & 0.00119 \\
        & DiffC & $t591$ & 0.004700 & 0.2230 & 0.7597 & 108.57 & 0.00277 \\
        & DiffC & $t498$ & 0.006910 & 0.1895 & 0.8399 & 89.98 & 0.00045 \\
        \midrule
        \multirow{10}{*}{MSCOCO30k}
        & \textbf{SAVAR} & 4 retained scales & \textbf{0.00295} & 0.2653 & 0.6370 & 54.41 & 0.03172 \\
        & \textbf{SAVAR} & 5 retained scales & \textbf{0.00460} & 0.1839 & 0.8422 & 4.21 & 0.00090 \\
        & ARPC & 4 retained scales & 0.00394 & 0.3336 & 0.5757 & 84.73 & 0.05367 \\
        & ARPC & 5 retained scales & 0.00684 & 0.2524 & 0.6581 & 58.80 & 0.04678 \\
        & PerCo & low & 0.00364 & 0.2452 & 0.8084 & 5.24 & 0.00203 \\
        & PerCo & mid & 0.03293 & 0.1643 & 0.9050 & 3.72 & 0.00150 \\
        & DLF & $q0$ & 0.009337 & 0.1533 & 0.8890 & 4.49 & 0.00150 \\
        & DLF & $q3$ & 0.017400 & 0.1299 & 0.9220 & 2.62 & 0.00049 \\
        & DiffC & $t591$ & 0.00470 & 0.2214 & 0.7133 & 12.87 & 0.00716 \\
        &         DiffC & $t498$ & 0.00691 & 0.1889 & 0.8023 & 6.57 & 0.00303 \\
        \bottomrule
    \end{tabular}
    }
\end{table*}

\section{Bitrate Accounting}
\label{app:bitrate_accounting}

SAVAR transmits two forms of coded information: retained VAR scale tokens and a caption. Both are included in every reported token-plus-caption bitrate. For a source image of size $H \times W$, the reported rate is
\begin{equation}
    R_{\mathrm{bpp}} =
    \frac{
        \sum_{i=1}^{m} h_i w_i \lceil \log_2 V \rceil
        +
        8\,|\mathrm{UTF8}(T)|
    }{HW},
\end{equation}
where $m$ is the number of retained scales, $V=4096$ is the VAR vocabulary size, and $T$ is the encoder-generated caption. The decoder receives the caption string and retained tokens only; it does not receive a dense text or image embedding.

For $512 \times 512$ evaluation, the two reported SAVAR operating points retain 30 and 66 VAR tokens, corresponding to $0.001373$ and $0.003021$ token bpp, respectively. The total bpp varies with the measured caption byte length. On Kodak-24, the average caption payload is $0.001242$ bpp, yielding the totals in Table~\ref{tab:appendix_bitrate_kodak}. On the larger datasets, we report the measured dataset-level total bpp directly in the main tables because caption lengths vary across images.

\begin{table}[h]
    \caption{Kodak-24 SAVAR bitrate decomposition at $512 \times 512$. Both retained VAR tokens and caption bytes are included in the total bpp.}
    \label{tab:appendix_bitrate_kodak}
    \centering
    \begin{tabular}{l c c c c}
        \toprule
        Operating point & Token count & Token bpp & Caption bpp & Total bpp \\
        \midrule
        4 retained scales & 30 & 0.001373 & 0.001242 & 0.002615 \\
        5 retained scales & 66 & 0.003021 & 0.001242 & 0.004263 \\
        \bottomrule
    \end{tabular}
\end{table}

\section{Metric and Evaluation Protocol}
\label{app:evaluation_protocol}

All images are evaluated at $512 \times 512$ resolution using the same preprocessing across Kodak, CLIC2020, and MSCOCO30k: short-side resizing to 512 pixels with LANCZOS interpolation, followed by a $512 \times 512$ center crop. We report PSNR and MS-SSIM for pixel-level fidelity, LPIPS and DISTS for perceptual distortion, FID and KID for distribution-level realism, and CLIP-I for source-image semantic similarity. All FID and KID values use the same Inception-v3 pool3 feature extractor and identical real/reconstruction preprocessing. FID and KID curves and the complete appendix table focus on CLIC2020 and MSCOCO30k, where aggregate distribution metrics are available across methods. These aggregate realism metrics are used only as supporting distribution-level measurements and should not be interpreted as primary measurements of per-image reconstruction fidelity. Kodak distribution-level values, when available in the main table, are treated as auxiliary because Kodak contains only 24 images and does not support the same complete aggregate comparison.

CLIP-I is the cosine similarity between image embeddings of the source and reconstructed images. Because CLIP-I depends on the fixed evaluator used for a given dataset, we use it only for comparisons within the same dataset and protocol. KID uses an unbiased estimator, so values near zero can be slightly negative.

For SAVAR, each reported operating point is evaluated on the reconstruction set produced by the fixed evaluation configuration for that operating point. The token-plus-caption bitrate is determined solely by the transmitted scale tokens and caption bytes. Baseline operating points are evaluated under the same $512 \times 512$ crop protocol. ARPC is reproduced from the authors' released code and checkpoints at the matched retained-scale counts. For PerCo, MS-ILLM, DiffC, and DLF, we use author-released reconstruction files when available; otherwise we reproduce the operating point with the authors' released inference code or checkpoints. We do not train any baseline model.

\section{Auxiliary BD-Rate Protocol for Table~\ref{tab:latency}}
\label{app:bdrate_protocol}

An auxiliary BD-Rate summary on Kodak-24 at $512 \times 512$ using DISTS as the distortion axis and total bpp as the rate axis is reported in Table~\ref{tab:latency}. MS-ILLM is the anchor curve. We use the Bjontegaard log-rate formulation~\cite{bjontegaard2001calculation}: each curve is represented as $\log R(D)$, where $R$ is bpp and $D$ is DISTS, and the reported value is
\begin{align}
    \mathrm{BD}\mbox{-}\mathrm{Rate}
    &=
    \Bigg(
    \exp\Bigg[
    \frac{1}{D_{\max}-D_{\min}}
    \int_{D_{\min}}^{D_{\max}}
    \Delta(D)\,dD
    \Bigg]
    - 1
    \Bigg) \times 100\%, \\
    \Delta(D)
    &=\log R_{\mathrm{method}}(D)-\log R_{\mathrm{MS}\mbox{-}\mathrm{ILLM}}(D).
\end{align}
The integration interval $[D_{\min},D_{\max}]$ is the overlap of the method curve and the MS-ILLM anchor in DISTS space; we do not extrapolate beyond the available points. Curves are sorted by DISTS before interpolation. For curves with more than two valid points, we use the standard cubic Bjontegaard interpolation in $\log R$ over DISTS; for two-point curves, this reduces to linear interpolation in $\log R$.

The auxiliary BD-Rate calculation uses the following Kodak DISTS-rate points. The main comparison tables retain only the target extreme-low bitrate rows, but the BD-Rate curve fit includes adjacent above-$0.01$ context points when a method provides them; these points are used only to define the local rate-distortion curve and are not included as extreme-low bitrate table rows. SAVAR and ARPC use only their two reported retained-scale points.

\begin{table*}[t]
    \caption{Kodak-24 DISTS-rate points used for the auxiliary BD-Rate calculation in Table~\ref{tab:latency}. Rates are total bpp. Points above $0.01$ bpp are included only as adjacent curve context for interpolation.}
    \label{tab:appendix_bdrate_points}
    \centering
    \resizebox{\textwidth}{!}{
    \begin{tabular}{l l l}
        \toprule
        Method & Operating points used & Notes \\
        \midrule
        MS-ILLM & $(0.007899,0.3392)$, $(0.011314,0.2867)$, $(0.0476,0.1512)$, $(0.0845,0.1233)$, $(0.1586,0.0937)$ & Anchor curve \\
        PerCo & $(0.003630,0.2421)$, $(0.032930,0.1645)$ & Includes the adjacent above-$0.01$ bpp point \\
        DLF & $(0.008928,0.1602)$, $(0.016925,0.1399)$ & Includes the adjacent above-$0.01$ bpp point \\
        DiffC & $(0.004684,0.2278)$, $(0.007014,0.1990)$ & Both points are below $0.01$ bpp \\
        ARPC & $(0.002710,0.2780)$, $(0.005690,0.2130)$ & Matched retained-scale points \\
        SAVAR & $(0.002615,0.2190)$, $(0.004263,0.1867)$ & Token-plus-caption total bpp \\
        \bottomrule
    \end{tabular}
    }
\end{table*}

\section{Broader Impacts and Asset Licenses}
\label{app:broader_impacts_licenses}

\textbf{Broader impacts.}
SAVAR targets efficient image transmission in the extreme-low bitrate setting, which could reduce bandwidth and storage costs for image communication, archival, and low-connectivity applications. As with other generative codecs, reconstructions are intended for bandwidth-efficient visual communication rather than forensic or diagnostic record keeping in sensitive settings. We mitigate this risk in the paper by explicitly reporting paired fidelity metrics and caption and token bitrate accounting.

\textbf{Existing assets and licenses.}
The work uses public research assets and released pretrained components under their stated terms. Evaluation uses Kodak PhotoCD images from the public Kodak test set, the CLIC2020 validation set under the CLIC challenge terms, and MSCOCO images under the licenses recorded by COCO/Flickr metadata; training uses ImageNet ILSVRC under its non-commercial research access terms. The captioner \texttt{Salesforce/blip2-opt-2.7b} and the \texttt{FoundationVision/VAR} code/checkpoints are released under MIT licenses, while \texttt{jinaai/jina-clip-v2} is released under CC-BY-NC-4.0. Baseline code, checkpoints, or reconstructions are used from the authors' released assets when available, including ARPC, PerCo, MS-ILLM, DiffC, and DLF, following the license or access terms stated by each repository, model release, or reconstruction package. We use these assets only for non-commercial research evaluation and cite the corresponding papers in the main text.

\section{Reproducibility Details}
\label{app:reproducibility_details}

\textbf{Architecture and training stages.}
We use a frozen VAR VQ-VAE with vocabulary size $V=4096$, latent channels $C_v=32$, base channel width 160, and multi-scale patch sizes $(1,2,3,4,6,9,13,18,24,32)$ for $512 \times 512$ images. The operating point determines how many scales are retained, and the partial VQ-VAE latent reconstruction from those scales defines $a_{\mathrm{src}}$. The SAVAR autoregressive backbone has depth 36, width $36 \times 64=2304$, 36 attention heads, and per-block AdaLN conditioning. Training proceeds in two stages: CAM is first trained as a standalone alignment module, and the SAVAR autoregressive backbone is then fine-tuned with the VQ-VAE, Jina-CLIP-v2, and CAM frozen.

\textbf{CAM pretraining.}
CAM is trained on ImageNet images paired with BLIP-2-generated captions, using 8 NVIDIA L20X GPUs and a global batch size of 1024. The contrastive term in $\mathcal{L}_{\mathrm{CAM}}$ applies cross-entropy in both prediction-to-target and target-to-prediction directions with a learnable logit scale, and its weight $w(t)$ is annealed from 0.04 to 0.004 over the first 6 epochs. We select the checkpoint with the lowest validation pairwise loss. No RGB reconstruction, LPIPS, DISTS, or perceptual loss is used in this CAM stage.

\textbf{Fine-tuning data and training schedule.}
We fine-tune the VAR-d36 autoregressive Transformer on the ImageNet ILSVRC train split and use the ImageNet validation split for model selection and monitoring. ImageNet provides the source images; the text condition is not taken from dataset annotations. Instead, each train and validation image is paired with a BLIP-2 caption generated from that image using the captioning setup below. Images use the same $512 \times 512$ short-side-resize and center-crop preprocessing as evaluation. The main extreme-low bitrate fine-tuning run uses the first five token scales, with patch sizes $(1,2,3,4,6)$ and 66 retained tokens for the five-scale operating point. We train for 15 epochs with global batch size 512. The training log contains 2483 optimizer steps per epoch, corresponding to approximately 37,245 optimizer steps. The VQ-VAE, image decoder, Jina-CLIP-v2 encoders, and CAM are frozen during this backbone fine-tuning stage; the autoregressive Transformer, token-based initialization projections, and AdaLN conditioning projections are updated.

\begin{table}[!ht]
    \caption{Fine-tuning hyperparameters for the SAVAR VAR-d36 backbone.}
    \label{tab:appendix_training_details}
    \centering
    \begin{tabular}{l l}
        \toprule
        Item & Setting \\
        \midrule
        Image split & ImageNet ILSVRC train / val \\
        Text source & BLIP-2 captions generated from each image \\
        Image preprocessing & Short-side resize and center crop to $512 \times 512$ \\
        Epochs / steps & 15 epochs / $\approx$37,245 optimizer steps \\
        Global batch size & 512 \\
        Optimizer & AdamW, betas $(0.9, 0.95)$, fused implementation \\
        LR schedule & Linear schedule; table/base LR $1{\times}10^{-5}$; Transformer LR $2{\times}10^{-5}$ \\
        Warmup / final LR & 2-epoch warmup from 0.5\% of LR; ending LR ratio 0.01 \\
        Differential LR scales & 0.1 for base groups; 20.0 for new condition/noise projection groups \\
        Weight decay & 0.05 for decay groups; no decay for no-decay groups \\
        Gradient clipping & 2.0 \\
        Condition dropout & 0.0 \\
        Other dropout & dropout 0.0, attention dropout 0.0, drop path 0.15 \\
        Random seed & 42 \\
        \bottomrule
    \end{tabular}
\end{table}

\textbf{Model size.}
The frozen VAR VQ-VAE has 108.95M parameters, split into a 44.11M encoder, 64.65M decoder, and 0.17M quantizer; component counts are rounded, so subtotals may differ slightly from the total. The VAR-d36 autoregressive backbone has 2.431B parameters. The combined VQ-VAE plus autoregressive model has 2.540B parameters. The backbone uses depth 36, width 2304, 36 attention heads, MLP ratio 4.0, vocabulary size 4096, and patch scales $(1,2,3,4,6,9,13,18,24,32)$ at $512 \times 512$ resolution.

\textbf{Decoding protocol.}
Unless otherwise stated, decoding uses classifier-free guidance scale 0.0, top-$k=900$, nucleus top-$p=0.95$, no smoothing, bfloat16 autoregressive precision, and a fixed pseudorandom seed specified in the implementation. The implementation does not apply an explicit temperature rescaling before token generation; equivalently, temperature is 1.0 after top-$k$/top-$p$ masking. Under this protocol, the decoder needs only the transmitted scale tokens and caption.

\textbf{Caption generation and text prompt.}
Text conditions are 1024-dimensional embeddings from \texttt{jinaai/jina-clip-v2}. The raw caption string is encoded directly; no additional prompt template is prepended or appended. For encoder timing and manifest caption generation, we use BLIP-2 \texttt{Salesforce/blip2-opt-2.7b} in image-only captioning mode with \texttt{max\_new\_tokens=40}. The Kodak timing measurement uses batch size 1 on one NVIDIA L20X GPU and reports an average BLIP-2 caption-generation time of 0.112 s per image; together with 0.034 s for VAR token encoding, this gives the 0.146 s total encoder-side runtime reported in Table~\ref{tab:latency}. The captioner runs only at the encoder and is not part of decoder latency.

\end{document}